%% file: acl2020.tex
\documentclass[11pt,a4paper]{article}
\usepackage[hyperref]{acl2020}
\usepackage{times}
\usepackage{latexsym}
\usepackage{graphicx}
\usepackage{amsmath}
\usepackage{amssymb}
\usepackage{subcaption}
\usepackage{multirow} 

\usepackage{xspace}
\usepackage[flushleft]{threeparttable}
\usepackage{soul}
\usepackage{multicol}
\usepackage{makecell}
\usepackage{array}

\usepackage{mathrsfs}
\usepackage{paralist} 
\usepackage{amsfonts}
\usepackage{booktabs}
\usepackage{tabu}
\usepackage{mathtools}
\usepackage{tabularx}

\usepackage{xcolor}

\usepackage{relsize}
\usepackage{graphicx}
\usepackage{tabularx}

\usepackage{footnote}
\makesavenoteenv{tabular}

\definecolor{asparagus}{rgb}{0.53, 0.66, 0.42}
\definecolor{americanrose}{rgb}{1.0, 0.01, 0.24}
\definecolor{ao}{rgb}{0.0, 0.5, 0.0}

\usepackage{microtype}

\aclfinalcopy 

\input{sections/macro}

\newcommand{\Ours}{\textsc{Photon}\xspace}

\newcommand{\spiderut}{$\text{Spider}_{\text{UTran}}$\xspace}
\newcommand{\texttosql}{text-to-SQL\xspace}

\title{\Ours: A Robust Cross-Domain Text-to-SQL System}

\author{
Jichuan Zeng$^{1}$\thanks{~~Equal contribution.
}~\thanks{~~This research was conducted during the author's internship at Salesforce Research.}
\quad Xi Victoria Lin$^{2}$\footnotemark[1]
\quad \textbf{Caiming Xiong}$^{2}$ 
\quad \textbf{Richard Socher}$^{2}$ \\
\quad \textbf{Michael R. Lyu}$^1$
\quad \textbf{Irwin King}$^1$ 
\quad \textbf{Steven C.H. Hoi}$^{2}$ \\
\affaddr{$^1$ The Chinese University of Hong Kong} \\
\affaddr{$^2$ Salesforce Research} \\
$^1$ \email{\{jczeng,lyu,king\}@cse.cuhk.edu.hk}\\
$^2$ \email{\{xilin,cxiong,rsocher,shoi\}@salesforce.com}
}

\date{}

\begin{document}

\maketitle

\input{sections/abstract}
\input{sections/intro}

\input{sections/system}
\input{sections/model}
\input{sections/evaluation}

\input{sections/related}
\input{sections/conclusion}


\bibliography{anthology,acl2020}
\bibliographystyle{acl_natbib}

\appendix
\input{sections/appendix}

\end{document}

%% file: sections/macro.tex
\newcommand{\bb}{\mathbf{b}}
\newcommand{\bff}{\mathbf{f}}

\newcommand{\bh}{\mathbf{h}}

\newcommand{\bs}{\mathbf{s}}
\newcommand{\be}{\mathbf{e}}

\newcommand{\bW}{\mathbf{W}}

\newcommand{\bzero}{\mathbf{0}}

\newcommand{\rom}[1]{%
	\textup{\uppercase\expandafter{\romannumeral#1}}%
}

\newcommand{\eat}[1]{\ignorespaces}

\newcolumntype{L}[1]{>{\raggedright\let\newline\\\arraybackslash\hspace{0pt}}m{#1}}
\newcolumntype{C}[1]{>{\centering\let\newline\\\arraybackslash\hspace{0pt}}m{#1}}
\newcolumntype{R}[1]{>{\raggedleft\let\newline\\\arraybackslash\hspace{0pt}}m{#1}}

\newcommand{\mathid}[1]{\ensuremath{\mathit{#1}}}
\def\|#1|{\mathid{#1}}
\protected\def\codeid#1{\ifmmode{\mbox{\smaller\ttfamily{#1}}}\else{\smaller\ttfamily
		#1}\fi}
\def\<#1>{\codeid{#1}}

\newcommand*{\affaddr}[1]{#1} 

\newcommand*{\email}[1]{\texttt{#1}}
\newcommand{\hide}[1]{}

%% file: sections/abstract.tex
\begin{abstract}
Natural language interfaces to databases (NLIDB) democratize end user access to relational data. 
Due to fundamental differences between natural language communication and programming, it is common for end users to issue questions that are ambiguous to the system or fall outside the semantic scope of its underlying query language.
We present \Ours, a robust, modular, cross-domain NLIDB that can flag natural language input to which a SQL mapping cannot be immediately determined.
\Ours consists of a strong neural semantic parser (63.2\% structure accuracy on the Spider dev benchmark), a human-in-the-loop question corrector, a SQL executor and a response generator. 
The question corrector is a discriminative neural sequence editor which detects confusion span(s) in the input question and suggests rephrasing until a translatable input is given by the user or a maximum number of iterations are conducted. 
Experiments on simulated data show that the proposed method effectively improves the robustness of text-to-SQL system against untranslatable user input. The live demo of our system is available at \url{http://naturalsql.com/}. 

\end{abstract}

%% file: sections/intro.tex
\section{Introduction}
\label{sec:intro}

Natural language interfaces to databases~\cite{DBLP:conf/iui/PopescuEK03,DBLP:conf/sigmod/LiJ14} democratize end user access to relational data and have attracted significant research attention for decades~\cite{DBLP:conf/naacl/HemphillGD90,DBLP:conf/naacl/DahlBBFHPPRS94,DBLP:conf/aaai/ZelleM96,DBLP:conf/iui/PopescuEK03,bertomeu-etal-2006-contextual,DBLP:journals/corr/abs-1709-00103,DBLP:conf/emnlp/YuZYYWLMLYRZR18,DBLP:journals/corr/abs-1909-05378}. 
Most existing NLIDBs adopt a modular architecture consisting of rule-based natural language parsing, ambiguity detection and pragmatics modeling~\cite{DBLP:conf/sigmod/LiJ14,DBLP:conf/uist/SetlurBTGC16,DBLP:conf/iui/SetlurTD19}. While they have been shown effective in pilot study and production, 
rule-based approaches are limited in terms of coverage, scalability and naturalness -- they are not robust against the diversity of human language expressions and are difficult to scale across domains. 

\begin{figure}[t]
	\centering
	\includegraphics[width=.48\textwidth]{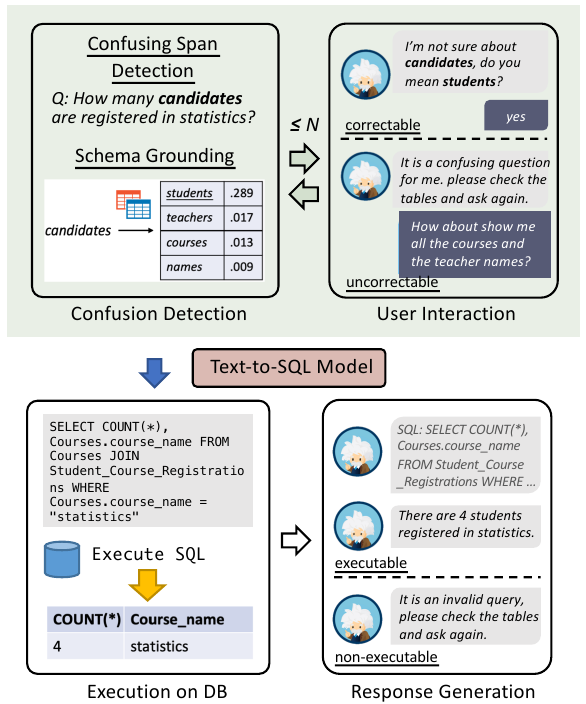}
	\caption{\Ours workflow. The question corrector (upper block) detects the untranslatable questions from user input, scans the confusion span(s) that need clarification or correction. The accepted question is  mapped into a SQL query through a text-to-SQL model, and finally the SQL execution results are returned to the user.}
	\label{fig:flow}
\end{figure}

\hide{
\begin{figure*}[t]
	\centering
	\includegraphics[width=.98\textwidth]{figures/workflow-cropped.pdf}
	\caption{Workflow of Photon. * is partially implemented in this version.}
	\label{fig:flow}
\end{figure*}
}

Recent advances in neural natural language processing~\cite{DBLP:conf/nips/SutskeverVL14,DBLP:conf/acl/DongL16,DBLP:conf/acl/SeeLM17,DBLP:conf/acl/LiangBLFL17,DBLP:journals/corr/abs-1905-13326,DBLP:conf/acl/BoginBG19}, pre-training~\cite{DBLP:conf/naacl/DevlinCLT19,DBLP:journals/corr/abs-1902-01069}, and the availability of large-scale supervised datasets~\cite{DBLP:journals/corr/abs-1709-00103,DBLP:conf/acl/RadevKZZFRS18,DBLP:conf/emnlp/YuZYYWLMLYRZR18,DBLP:conf/acl/YuZYTLLELPCJDPS19,DBLP:journals/corr/abs-1909-05378} enabled deep learning based approaches to significantly improve the state-of-the-art in nearly all subtasks of building an NLIDB. These include semantic parsing~\cite{DBLP:conf/acl/LapataD18,DBLP:conf/emnlp/YuYYZWLR19}, ambiguity detection and confidence estimation~\cite{DBLP:conf/acl/QuirkLD18,DBLP:conf/emnlp/YaoSSY19}, natural language response generation~\cite{DBLP:conf/acl/LiuLYWCS19}
and so on. Moreover, by jointly modeling the natural language question and database schema in the neural space, latest text-to-SQL semantic parsers can work cross domains~\cite{DBLP:conf/emnlp/YuZYYWLMLYRZR18,DBLP:conf/emnlp/YuYYZWLR19}.

We present \Ours, a modular, cross-domain NLIDB that adopts deep learning in its core components. \Ours consists of (1) a neural semantic parser, (2) a human-in-the-loop question corrector, (3) a SQL query executor and (4) a natural language response generator. The \emph{neural semantic parser} assumes limited DB content access due to data privacy concerns (\S~\ref{sec:semantic-parser}). It employs a BERT-based~\cite{DBLP:conf/naacl/DevlinCLT19} DB schema-aware question encoder and a pointer-generator decoder~\cite{DBLP:conf/acl/SeeLM17} with static SQL correctness check. It achieves competitive performance on the popular cross-domain text-to-SQL benchmark, Spider~\cite{DBLP:conf/emnlp/YuZYYWLMLYRZR18} (63.2\% structure accuracy on the dev set based on the official evaluation).\footnote{We are continuously improving the performance of the neural semantic parser. Currently the semantic parser only accepts standalone question as input. We plan to also model the interaction context in future work.} The \emph{question corrector} is a neural sequence editor which detects potential confusion span(s) in the input question and suggests possible corrections for the user to give feedback. When an input question is successfully translated into an executable SQL query, the \emph{response generator} generates a natural language response conditioned on the output of the SQL \emph{query executor}. 

A pilot study with non-expert SQL users shows that the system effectively increases the flexibility of user's natural language expression and is easy to be adapted to unseen databases. Being able to detect and correct untranslatable questions reduces unexpected error cases during user interaction.

%% file: sections/system.tex
\section{System Design}
\label{sec:system}

In this section, we will elaborate on the system design of \Ours.

\subsection{Overview}
Figure~\ref{fig:flow} shows the overall workflow of our system. \Ours is an end-to-end system that takes a user question and database schema as input, and output the query result after executing the generated SQL on the database.
\Ours is a modular framework designed towards practical industrial applications. The core modules in \Ours are the SQL parser and confusion detection mechanism. 
The SQL parser parses the input question and database schema, maps them into executable SQL query via an encoder-decoder framework. 
The confusion detection module identifies the untranslatable questions and captures the confusing span of the untranslatable question. The confusing tokens together with the context are fed into the auto-correction module to make a prediction of user attempted question.

To make it more applicable and accessible for user to query the database in a natural way, \Ours also provides user interaction module enabling user to refine their queries in the interaction with the system. 
Response generation handles the output of the system by transducing the database-style query result into natural language or post warning when the query is non-executable on the database, making the system more user-friendly. Notice that the response generation module in the current version is implemented using a template-based approach and can be improved by using more advanced response generation models.

\begin{figure}[htpb]
	\centering
	\includegraphics[width=.48\textwidth]{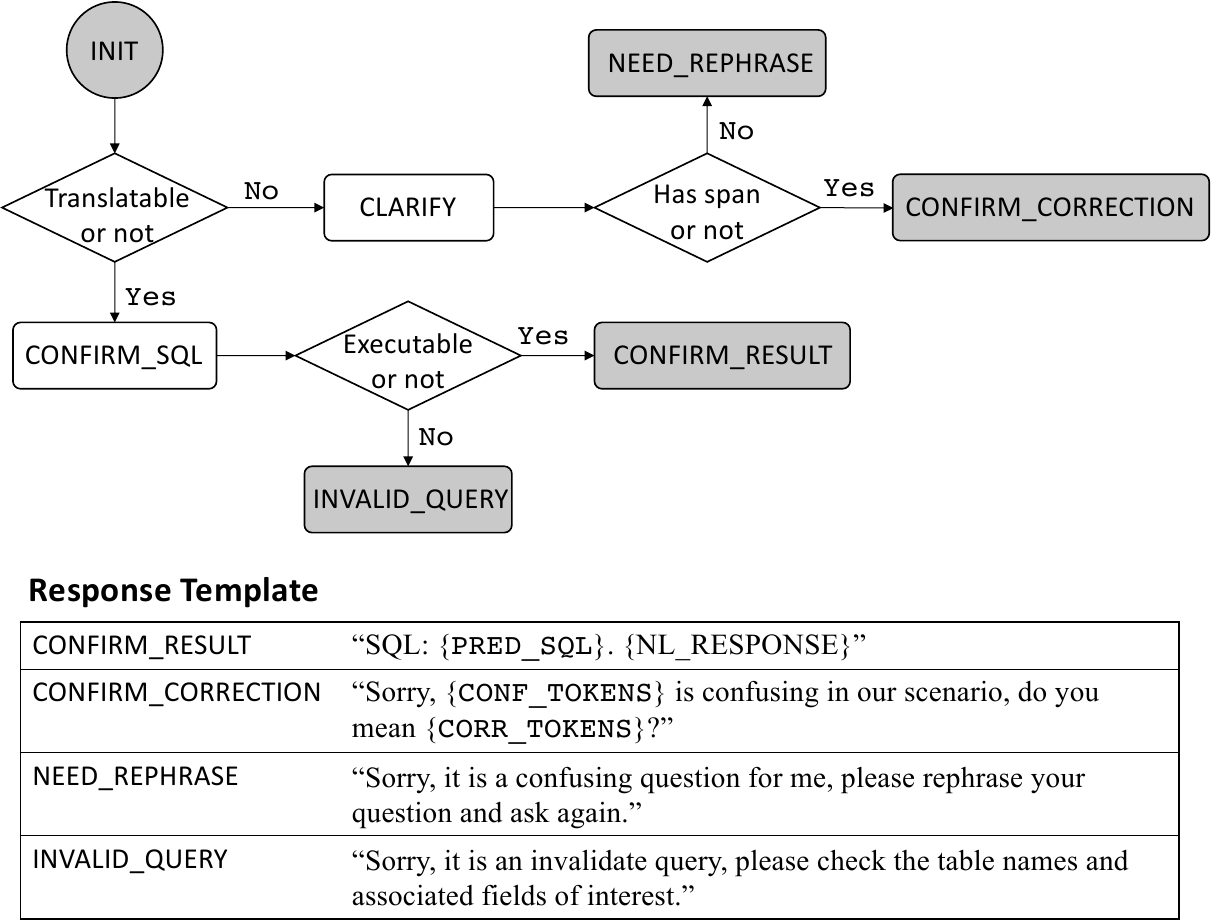}
	\caption{State transition map of interaction in \Ours. States with darker background are the end states that can receive user reply, and switch to \texttt{INIT} state automatically. The bottom part is the system response templates in each end state.}
	\label{fig:states}
\end{figure}

\begin{figure*}[t]
	\centering
	\includegraphics[width=.99\textwidth]{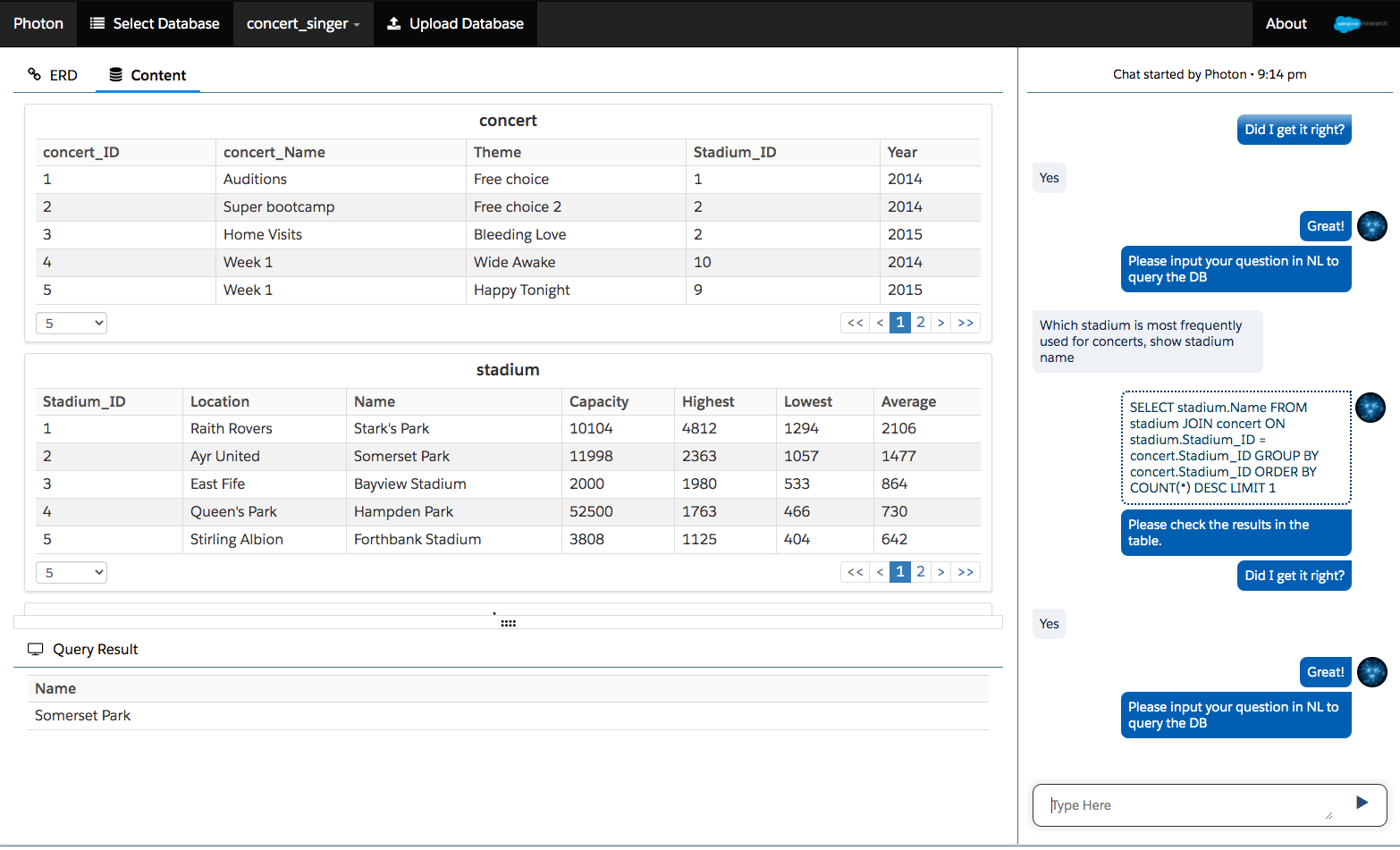}
		\vspace{-0.1in}
	\caption{\Ours main interface.}
	\label{fig:interface}
\end{figure*}

\subsection{User Interaction}

Figure~\ref{fig:states} illustrates the interaction process, which involves four types of response states: \texttt{CONFIRM\_RESULT}, \texttt{CONFIRM\_CORRECTION}, \texttt{NEED\_REPHRASE}, and \texttt{INVALID\_QUERY}. The set of response templates can be found at the bottom of Figure~\ref{fig:states}. When a user initiates the conversation by entering one query, \Ours will first predict whether the query is translatable or not. If translatable, \Ours generates the corresponding SQL command and checks the command's executability; otherwise, \Ours will provide a correction strategy (i.e., \texttt{CONFIRM\_CORRECTION}) based on the detected confusing span or ask the users to further rephrase the inquiry (i.e., \texttt{NEED\_REPHRASE}) if no span is captured.

\subsection{UI Design}
Our system UI consists of three panels: chat window, schema viewer and results viewer (Figure~\ref{fig:interface}). 
\begin{itemize}
    \item \emph{Chat window}: This is a standard chat window that facilitates communication between the user and \Ours. The user types the natural language input and the natural language responses of the system are displayed. 
    \item \emph{Schema viewer}: This view provides a graph visualization of the underlying relational DB schema. The panel is hideable and will not be shown in case the DB schema is confidential.
    \item \emph{Result viewer}: This view displays the returned results of an executable SQL query mapped from a confirmed input question. 
    The SQL query is formatted and displayed in the top for user verification.
    Multi-record results are presented as sub-tables. Result consists of a single table cell is presented as a 1-cell sub-table. If the result comes from an aggregation operation such as a counting, the data records supporting the calculation are also shown for explanability. Confidential DB records are hidden from the display and the user is informed of the number of hidden records. 
\end{itemize}

\subsection{Cross Domain} A relational DB for user queries should be set before usage. \Ours consists of a collection of default databases and allows users to upload their own DBs for testing. Users can select which database they want to query by clicking the ``Selected Database'' drop down button. 

\subsection{Dual Query Mode} 
\Ours accepts both natural language questions and well-formed SQL queries as input. It automatically detects the user input type and executes the input immediately if it is a valid SQL query. We observed that the dual query model can be more efficient than NL-only mode, especially for users who have SQL background.


%% file: sections/model.tex
\section{Model}

\hide{
\subsection{Preprocessing}
After taking the user question and some database schema as the input, 
\Ours first employs WordPiece tokenizer~\cite{DBLP:journals/corr/WuSCLNMKCGMKSJL16} from BERT~\cite{DBLP:conf/naacl/DevlinCLT19} to the get tokens of the user question. 
Database schema is the skeleton of a database, including the structural information of tables and columns, such as the table/column names, column types, primary keys, and foreign keys. We also use WordPiece tokenizer for the table/column names and represent the primary keys, foreign keys, and column types as discrete numbers.
}

\subsection{Neural Semantic Parser}
\label{sec:semantic-parser}

\begin{figure*}[t]
	\centering
	\includegraphics[width=.95\textwidth]{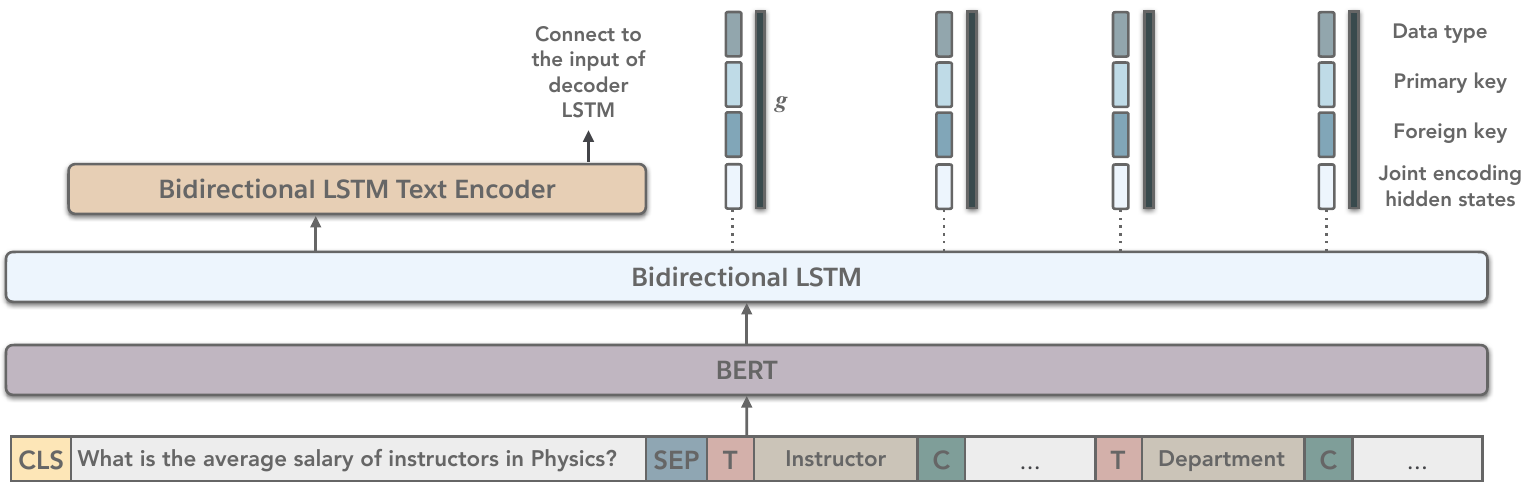}
		\vspace{-0.1in}
	\caption{Joint schema-question encoder.}
	\label{fig:text2sql-encoder}
\end{figure*}

\begin{figure*}[t]
	\centering
	\includegraphics[width=\textwidth]{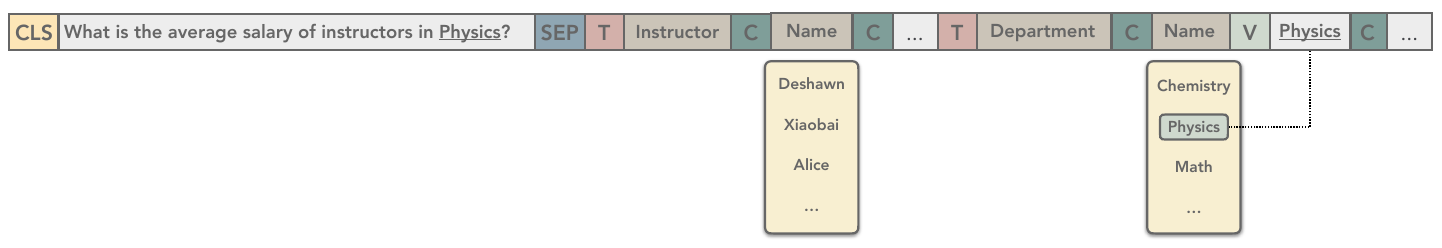}
	\vspace{-0.2in}
	\caption{Joint schema-question encoder with picklist value augmentation.} 
	\label{fig:text2sql-encoder-picklist}
\end{figure*}

The neural semantic parser is an end-to-end model whose input consists of a user question and the DB schema, and outputs a SQL query. Due to data privacy concerns, we assume that the neural semantic parser does not have full access to the DB content. Instead, we assume for each DB field, the parser have access to the set of possible values of the field, for example, ``Country.Region'': \{``Carribean'', ``Porto Rico'', ...\}\footnote{In practice, we can limit the access to only certain fields.}. We call such value sets ``picklists'' by industry convention.

\subsubsection{Schema-Question Encoder}

Following previous work~\cite{DBLP:journals/corr/abs-1902-01069,DBLP:conf/emnlp/YuYYZWLR19}, we serialize the relational DB schema and concatenate it to the user question. As shown in Figure~\ref{fig:text2sql-encoder} 
, we represent each table with the table name followed by a sequence of field names. Each table name is preceded by the special token \<[T]> and each field name is preceded by the special token \<[C]>. The representations of multiple tables are concatenated together to form the serialization of the schema, which is surrounded by \<[SEP]> tokens and concatenated to the question. Finally, the question is preceded by the \<[CLS]> token following convention of BERT encoder~\cite{DBLP:conf/naacl/DevlinCLT19}. 

This sequence is fed into the pretrained BERT, followed by a bi-directional LSTM to form a joint encoding of the question and schema $\bh_{\text{input}}$. The text portion of $\bh_{\text{input}}$ is passed through another bi-LSTM to obtain the question encoding $\bh_{\text{Q}}$. We represent each schema component (tables and fields) using the slices of $\bh_{\text{input}}$ corresponding to the special token \<[T]> and \<[C]>. 

\paragraph{Meta-data Features}We further trained dense look-up features to represent if a field is a primary key ($\bff_{\text{pri}}$), if a field appears in a foreign key pair ($\bff_{\text{for}}$) and the data type of the field ($\bff_{\text{type}}$). These meta-data features are fused with the representations in $\bh_{\text{input}}$ via a projection layer $g$ to obtain the final representation of each schema component:
\begin{align}
    \bh^{C_p} &= g([\bh_{\text{input}}^m; \bff_{\text{pri}}^i; \bff_{\text{for}}^j; \bff_{\text{type}}^k]) \\
              &= \text{ReLU}(\bW_g[\bh_{\text{input}}^m; \bff_{\text{pri}}^i; \bff_{\text{for}}^j; \bff_{\text{type}}^k] + \bb_g) \nonumber\\
    \bh^{T_q} &= g([\bh_{\text{input}}^n; \bzero; \bzero; \bzero]),
\end{align}
where $m$ is the index of the special token corresponding to the $p$-th column in the input and $n$ is the index of the special token corresponding to the $q$-th table in the input. $i$, $j$ and $k$ are the feature indices indicating the corresponding properties of $C_p$. $[\bh_{\text{input}}^m; \bff_{\text{pri}}^i; \bff_{\text{for}}^j; \bff_{\text{type}}^k]$ is the concatenation of the four vectors. The meta-data features we include are specific to fields and the table representations are fused with zero place-holder vectors.

\subsubsection{Decoder}
We use an LSTM-based sequential pointer-generator~\cite{see-etal-2017-get} as the decoder. The generation vocabulary of our decoder consists of 70 SQL keywords and reserved tokens, plus the 10 digits\footnote{Such that the parser is able to generate numbers corresponding to utterances such as ``first'', ``second'' etc.}. At each step, the decoder computes a probability distribution over actions that consists of generating a token from the reserved vocabulary, copying a token from the input text or copying a schema component.

\subsubsection{Static SQL Correctness Check}
The sequential pointer-generator we adopted does not guarantee the output SQL is syntactically correct. In practice, we perform beam-search decoding and run a static SQL correctness check\footnote{Some prior work such as~\citep{DBLP:journals/corr/abs-1807-03100} performs a similar check by executing the decoded SQL queries on the target DB. We implement the static checking as it can reduce the traffic between the interface and the DB.} to eliminate erroneous predictions from the beam. Specifically, we employ a tool implemented on top of the Mozilla SQL Parser\footnote{\url{https://github.com/mozilla/moz-sql-parser}} to analyze the output SQL queries and ensure they satisfy the following criteria:

\vspace{6px}
1. The SQL query is syntactically correct.

2. The SQL query satisfies schema consistency\footnote{We define the following schema consistency rules: the fields appeared in a \<SELECT> SQL query must come from the tables in the corresponding \<FROM> clause; the fields in a \<JOIN> condition clause must come from tables mentioned in front of them in the \<JOIN> clause.}.

\vspace{6px}
We found this approach very effective and results in an absolute improvement of 4$\sim$5\% in the evaluation score on Spider dev set~\cite{DBLP:conf/emnlp/YuZYYWLMLYRZR18}.

\subsubsection{Picklist Incorporation}

We use picklists to inform the semantic parser regarding potential matches in the DB. For an input question $Q$ and a field $C_p$, we compute the longest character sequence match between $Q$ and each value in the picklist of $C_p$. We select the value with top-1 matching score above a certain threshold $\theta$ as a match. For each field with a matched picklist value, we append the surface form of the value to it in the input sequence representation, separated by the special token \<[V]>. The augmented sequence is used as the input to the schema-question encoder. In practice, we found picklist augmentation results in an absolute performance improvement of 1\% on the Spider dev set. 

Figure~\ref{fig:text2sql-encoder-picklist} illustrates the input sequence with augmented picklist values. In this example, the matching algorithm identifies ``Carribean'' associated with the column ``Country.Region'' as a match. Hence it inserts ``Carribean'' after [... \<[C]>, ``Region''] with \<[V]> as a separation token\footnote{In practice, we found a question typically has 0 to 4 picklist value matches. As a result, the picklist augmented schema-question representation still stays under the maximum input length of BERT.}. 
The representations of fields with no picklist value match are unchanged.

\subsection{Confusion Detection: Handling Untranslatable and Ambiguous Input}

In order to handle ambiguous and untranslatable input questions, \Ours adopts a discriminatively trained classifier to detect user input to which a SQL mapping cannot be immediately determined. This covers questions that are incomplete (e.g. \emph{What is the total?}), ambiguous or vague (e.g. \emph{Show me homes with good schools}), beyond the representation scope of SQL (e.g. \emph{How many tourists visited all of the 10 attractions?}), or simply noisy (e.g. \emph{Cyrus teaches physics in department}). 
\hide{
Existing text-to-SQL datasets and models assume that all the questions can be translated in to the corresponding SQL queries, which is impractical in many real-world applications. Therefore, we utilize UTran-SQL, an adversarial text-to-SQL dataset consisting of untranslatable questions, to build the untranslatable question detection modular of \Ours. 
}

\subsubsection{Untranslatable Question Detection}

Inspired by~\cite{DBLP:conf/acl/RajpurkarJL18}, we create a synthetic dataset which consists of untranslatable questions generated by applying rule-based transformations and adversarial filtering~\cite{DBLP:conf/emnlp/ZellersBSC18} to examples in existing text-to-SQL datasets.
We then train a stagewise model that first classifies if the input is translatable or not, and then predicts confusing spans in an untranslatable input.
\hide{
Specifically, the set of modification rules consists of the \emph{Swap} and \emph{Drop} strategies from question-side and \emph{Drop} strategy from schema-side. Illustration examples are listed in Table~\ref{tab:transform_example}.
For example, question \emph{Drop} strategy first search the table column related tokens by n-gram matching, randomly remove one together with its proposition if applicable. 
Other strategies have a similar methodology and procedure.
}

\paragraph{Dataset Construction.} In order to construct the untranslatable questions, we firstly exam the types of untranslatable questions seen on the manually constructed CoSQL~\cite{DBLP:journals/corr/abs-1909-05378} and Multi-WOZ~\cite{DBLP:conf/emnlp/BudzianowskiWTC18} datasets (Table~\ref{tab:question-categorization} of~\ref{sec:app-ambiguous}). 
We then design our modification strategies to generate the untranslatable questions from the original text-to-SQL dataset automatically. 
Specifically, for a text-to-SQL example that contains a natural language question, a DB schema and a SQL query, we first identify all non-overlapping question spans that possibly refer to a table field occurred in the \texttt{SELECT} and \texttt{WHERE} clauses of the SQL query using string-matching heuristics. Then we apply \emph{Swap} and \emph{Drop} operations on the question and DB schema respectively to generate different types of untranslatable questions. 
The modification tokens are marked as the confusion spans of the synthetic untranslatable questions, except for the question \emph{Drop} strategy.

Table~\ref{tab:transform_example} in~\ref{sec:app-ambiguous}
provides a detailed summary of all transformations applied. To introduce semantic variation and ensure grammar fluency, we apply back-translation on the generated question using Google Cloud Translation API\footnote{ \url{https://cloud.google.com/translate/}. We use Chinese as the intermediate language.}. 
For example, given the original question ``How many countries exist?'', ``countries'' is detected to be referring to a table field. We drop the token, and pass the modified question ``How many exist?'' to back-translation for grammar smoothing. After that, we obtain the untranslatable question ``How many are there?''. 
Once we have the synthetic untranslatable questions, adversarial filtering is employed to iteratively refine the set of untranslatable examples to be more indiscernible by trivial stylistic classifiers~\cite{DBLP:conf/emnlp/ZellersBSC18}.


\paragraph{Predicting Untranslatable Questions and Confusing Spans.}

We utilize the BERT-based contextualized question-schema representations produced by the encoder in Figure~\ref{fig:text2sql-encoder-picklist} to extract the confusion span in a question hence predict its translatability. To do so, we normalize\footnote{Essentially, when the question is untranslatable but there is no identifiable confusion span, the whole question is marked as confusing; when the question is translatable, \<[CLS]> is used as the dummy confusion span.} the aforementioned synthetic dataset such that each question $Q$ in an example is paired with a span label $\zeta$, where
\begin{equation}
    \zeta = \begin{cases}
        (s+1, t+1),& \text{if } Q \text{ is untranslatable with }\;\;\\
        & \text{conf. span } (s, t)\\
        (1, |Q|),& \text{if } Q \text{ is untranslatable with }\;\;\\
        & \text{no identifiable conf. span}\\
        (0, 0).& \text{if } Q \text{ is translatable}
    \end{cases}
\end{equation}
We use the span extractor proposed by~\citet{DBLP:conf/naacl/DevlinCLT19} for SQuAD v1.1~\cite{Rajpurkar2016SQuAD10}. Specifically, the extractor introduces a start vector $\bs$ and an end vector $\be$. The probability of word $i$ being the start of the answer span is conputed as a dot product between $\bh_{\text{input}}^{i}$ and $\bs$ followed by a softmax over \<[CLS]> and all words in the question. The end of the answer span is predicted in a similar manner.
\begin{align}
    p_{\text{start}}^i=\frac{e^{\bs^{\top}{\bh_{\text{input}}^{i}}}}{\sum_j e^{\bs^{\top}{\bh_{\text{input}}^j}}},\  
    p_{\text{end}}^i=\frac{e^{\be^{\top}{\bh_{\text{input}}^{i}}}}{\sum_j e^{\be^{\top}{\bh_{\text{input}}^j}}}.
\end{align}
The score of a candidate span $(i, j)$ is $\bs^{\top}{\bh_{\text{input}}^{i}}+\be^{\top}{\bh_{\text{input}}^{j}}$, and the maximum scoring span with $j\geq i$ is used as a prediction. The training objective is the sum of the log-likelihood of the correct start and end indices.
\hide{
We utilize the BERT-based contextualized representations of \texttt{[CLS]} token, followed by a single-layer classifier to tell whether a given user question and table schema can be translated into SQL or not. 
To identify the questionable token spans of untranslatable question,
following~\citet{DBLP:conf/emnlp/YuYYZWLR19}, we employ a hierarchical bi-LSTM structure to encode each column header and use the hidden states as the column header embedding. We then use a bi-LSTM to encode the question's BERT embedding, and the hidden states are fed into a dot-product co-attention~\cite{DBLP:conf/emnlp/LuongPM15} layer over the column header embedding. The output of co-attention augmented question embedding is fed into a linear layer follow by softmax operator to predict the start and end tokens indices of the confusing spans in the question.
}

\subsubsection{Database-aware Token Correction}
\begin{figure}[htpb]
	\centering
	\includegraphics[width=.48\textwidth]{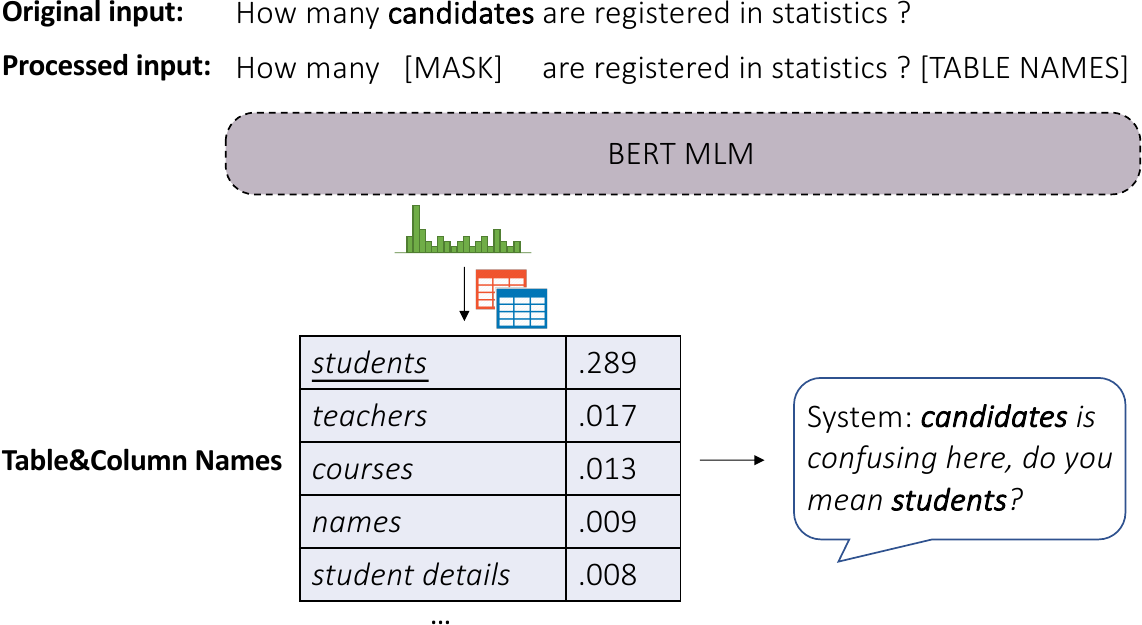}
	\vspace{-0.2in}
	\caption{Token Correction in \Ours.}
	\label{fig:corr}\vspace{-0.1in}
\end{figure}

Figure~\ref{fig:corr} illustrates the proposed tokens correction module in \Ours. We use the masked language model (MLM) of BERT~\cite{DBLP:conf/naacl/DevlinCLT19} to auto-correct the confusing tokens. Specifically, we replace the confusing tokens with the \texttt{[MASK]} special token. 
The output distribution of MLM head on the mask token is employed to score the candidate spans. We construct the candidate span list by extracting all the table names and columns names from the database schema. After user confirmation, the confusing tokens in the input are replaced by the predicted tokens of MLM.

\hide{
\begin{figure}[t]
	\centering
	\begin{subfigure}[h]{0.48\textwidth}
	\includegraphics[width=1 \textwidth]{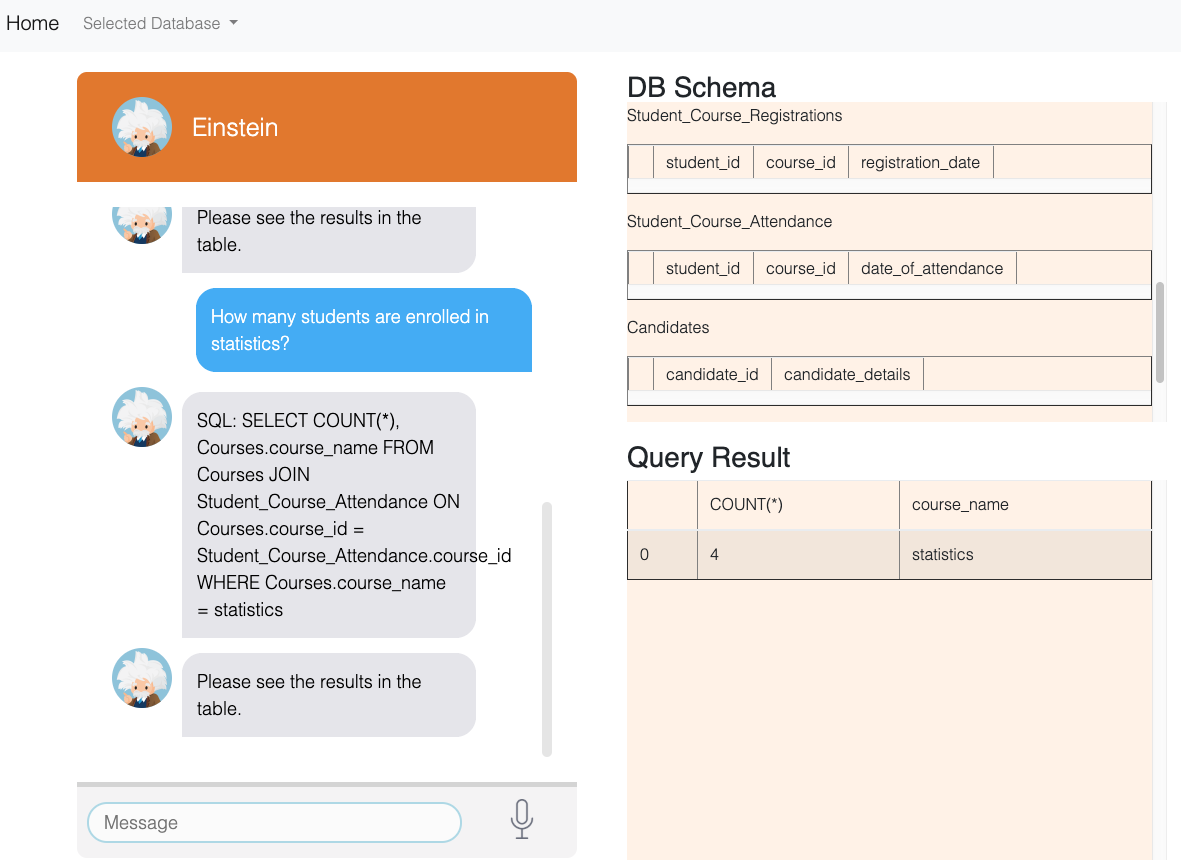}
    \caption{Translatable Case.}
    \end{subfigure}
    \hfill
    \begin{subfigure}[h]{0.48\textwidth}
    \includegraphics[width=1 \textwidth]{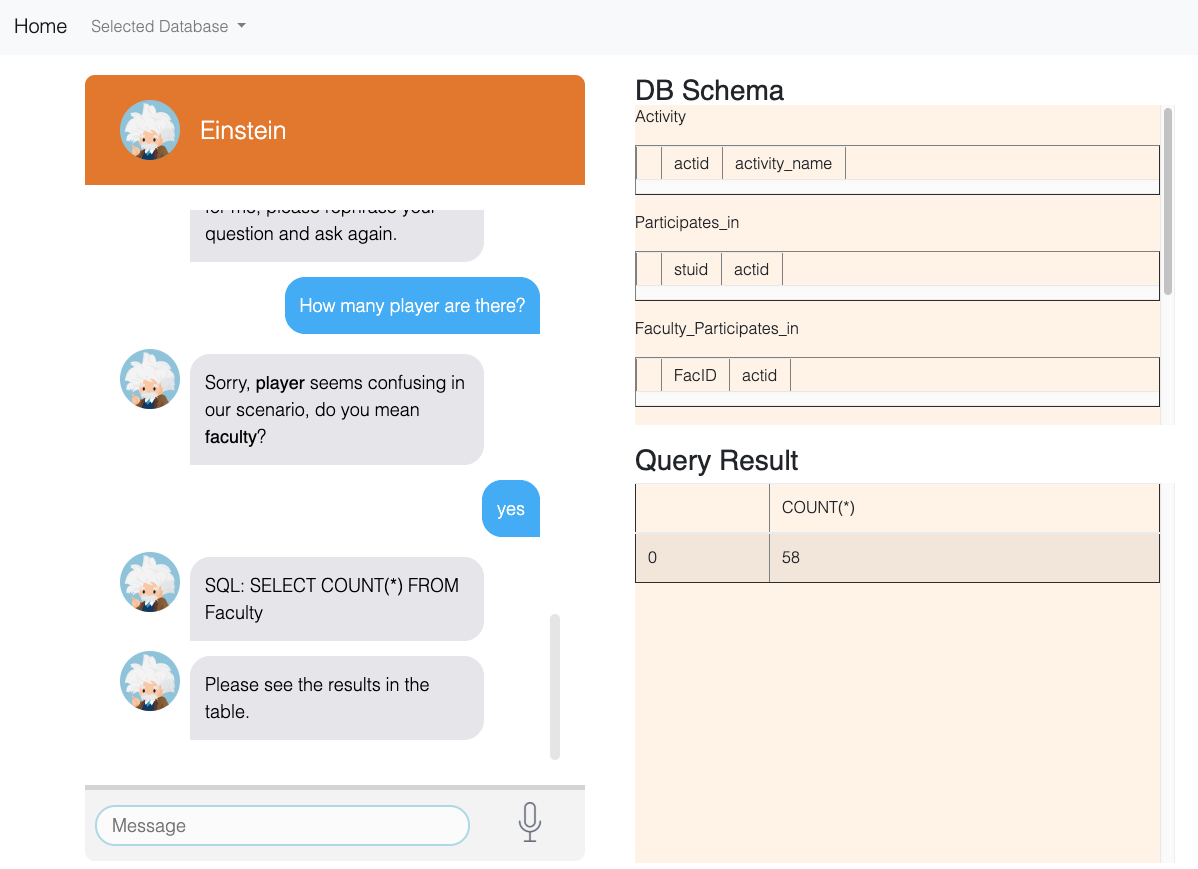}
    \caption{Untranslatable Case.}
    \end{subfigure}
    \centering
    \caption{
    The UI of \Ours.
    }
\label{fig:ui}
\end{figure}
}

%% file: sections/evaluation.tex
\section{Evaluation}
\label{sec:experiments}
In this section, we empirically evaluate the robustness and effectiveness of \Ours.

In particular, we examine two key modules of \Ours: the confusion detection module and the neural semantic parser. The former aims to detect the untranslatable questions and predicts the confusing spans; if the question is translatable, it then applies the proposed neural semantic parser to perform the text-to-SQL parsing. Since \Ours is designed as a stagewise system, we can evaluate the performance of each module separately. 


\subsection{Experimental Setup} 

\paragraph{Dataset.} 
We conduct experiments on Spider~\cite{DBLP:conf/emnlp/YuZYYWLMLYRZR18} and \spiderut dataset. Spider is a large-scale, human annotated, cross-domain text-to-SQL benchmark. \spiderut is our modified dataset to evaluate robustness, created by injecting the untranslatable questions into Spider. We obtained 5,330 additional untranslatable questions (4,733 for training and 597 for development) from the original Spider dataset. To ensure the quality of our synthetic dataset, we hired SQL experts from Upwork\footnote{\url{https://www.upwork.com/}} to annotate the auto-generated untranslatable examples in the dev set.
We conduct our evaluation by following the database split setting, as illustrated in Table~\ref{tab:dataset_split}. 
The split follows the original dataset hence there is no test set of \spiderut (the test set of Spider is not publicly accessible).

\begin{table}[htbp]
\centering
\scalebox{0.85}{
\setlength\tabcolsep{4.8pt}
\begin{tabular}{l|rr|rr}
\toprule
    & \multicolumn{2}{c|}{\bf Spider} & \multicolumn{2}{c}{\bf \spiderut} \\
    & Train & Dev & Train & Dev \\
     \midrule
\# Q & \quad 8,659\quad & \quad1,034\quad& \quad13,392\quad & \quad1,631\quad \\
\# UTran Q & 0 & 0 & 4,733 & 597 \\
\# Schema \quad\quad& 146 & 20 & 918 & 112 \\
\bottomrule
\end{tabular}
}
\caption{Data split of Spider and \spiderut. Q represents the all the questions, UTran Q represents the untranslatable questions.}
\label{tab:dataset_split}
\end{table}

\paragraph{Training and Inference Details.}
Our neural semantic parser is trained on Spider. We permute table order (up to 16 different ones) during training. We use the uncased BERT-base model from Huggingface's transformer library~\cite{Wolf2019HuggingFacesTS}. We set all LSTMs to 1-layer and set the dimension of $\bh_{\text{input}}$, $\bff_{\text{pri}}$, $\bff_{\text{for}}$, $\bff_{\text{type}}$ and the decoder to 512. We employ Adam-SGD~\cite{DBLP:journals/corr/KingmaB14} with a mini-batch size of 32 and default Adam parameters. We train a maximum of 50,000 steps and set the learning rate to $5e-4$ in the first 5,000 iterations and linearly decays it to 0 afterwards. We fine-tune BERT with a fine-tuning rate linearly increasing from $3e-5$ to $8e-5$ in the first 5,000 iterations and linearly decaying to 0 afterwards. We use a beam size of 128 in the beam search decoding.

Our confusion span extractor is trained on \spiderut. It uses the same encoder architecture as the semantic parser. We set the dimension of $\bs$ and $\be$ to 512. During inference, we determine a question as untranslatable if the predicted span start $s > 0$. \Ours highlights the confusion span to the user when it contains 5 or less tokens; otherwise, it generates a generic response prompting the user to rephrase.




\hide{
Different from the previous evaluation metric on Spider, without considering the value when computing the exact set match. We employ the complete exact set match accuracy including the values in SQL. Following UTran-SQL, for the untranslatable question, both correct translatability prediction and confusing span prediction receives 1 in accuracy, and 0 otherwise. }

\subsection{Experimental Results}

\paragraph{Confusion Detection.} We examine the robustness of \Ours by evaluating the performance of the confusion detection module in handling ambiguous and untranslatable input. In particular, we aim to examine if \Ours is effective in handling untranslatable questions by measuring its translatability detection accuracy and confusing span prediction accuracy \& F1 score\footnote{We use the same way as SQuAD 2.0~\cite{DBLP:conf/acl/RajpurkarJL18} to compute the span accuracy and F1.}. 
We compare to a baseline that uses a single-layer attentive bi-directional LSTM (``Att-biLSTM"). Table~\ref{table:sub_exp} shows the evaluation results on the \spiderut dataseet.

\begin{table}[htpb]
\center
\scalebox{0.85}{
\begin{tabular}{l|r|r|r}
\toprule
          &  {\bf \quad Tran Acc} &  {\bf Span Acc} & {\bf Span F1}\\ \midrule
Att-biLSTM & 66.6 & 58.7 & 59.2 \\
\Ours &  \textbf{87.6}  &  \textbf{83.6} & \textbf{84.7}  \\ \bottomrule
\end{tabular}}
\caption{Translatability prediction accuracy (``Tran Acc") and the confusing spans prediction accuracy and F1 on our \spiderut dataset (\%).}
\label{table:sub_exp}
\end{table}

As observed from Table~\ref{table:sub_exp}, \Ours achieves encouraging performance in determining the translatability of a question and predicting the confusing spans of untranslatable ones. In comparison to the Att-biLSTM baseline, \Ours obtains significant improvements in both translatability accuracy and the confusing spans prediction accuracy. These improvements are partly attribute to the proposed effective schema encoding strategy. 

\paragraph{Neural Semantic Parser.}

We then evaluate the performance of the proposed neural semantic parser of \Ours on the original Spider dataset. In particular, we compare \Ours and other existing text-to-SQL approaches by measuring the exact set match (EM) accuracy~\cite{DBLP:conf/emnlp/YuZYYWLMLYRZR18}. We compare with 
several existing approaches, including Global GNN~\cite{DBLP:journals/corr/abs-1908-11214}, 
EditSQL~\cite{DBLP:conf/emnlp/YuYYZWLR19}, IRNet~\cite{DBLP:conf/acl/GuoZGXLLZ19}, and RYANSQL~\cite{DBLP:journals/corr/abs-2004-03125}.
Table~\ref{table:spider_exp} shows the evaluation results on Spider Dev set.

\begin{table}[htpb]
    \centering
    \scalebox{0.85}{
    \begin{tabular}{l|c}
    \toprule
     {\bf Model}  &  {\bf EM Acc.} \\
 \midrule
    GNN~\cite{DBLP:conf/acl/BoginBG19} & 40.7 \\
    Global-GNN~\cite{DBLP:journals/corr/abs-1908-11214} & 52.7 \\
    EditSQL + BERT~\cite{DBLP:conf/emnlp/YuYYZWLR19}& 57.6 \\
    GNN+Bertrand-DR$^\dagger$~\cite{kelkar2020bertrand} & 57.9
    \\
    EditSQL+Bertrand-DR$^\dagger$~\cite{kelkar2020bertrand} & 58.5
    \\
    IRNet + BERT~\cite{DBLP:conf/acl/GuoZGXLLZ19} & 61.9 \\ 
    RYANSQL + BERT $^\dagger$ ~\cite{DBLP:journals/corr/abs-2004-03125}& 66.6 \\ \midrule
    \Ours & 63.2 \\ \bottomrule
    \end{tabular}
    }
    \begin{flushleft} {\quad\small $\dagger$ denotes unpublished work on arXiv.}\end{flushleft}
            \vspace{-0.1in}
    \caption{Experimental results on the Spider Dev set (\%). EM Acc. denotes the exact set match accuracy. 
    }
    \label{table:spider_exp}
\end{table}

As observed from Table~\ref{table:spider_exp}, \Ours achieves a very competitive text-to-SQL performance on the Spider benchmark with 63.2\% exact set match accuracy on the Spider dev set, which validates the effectiveness of our neural semantic parser for translating an input question into a valid SQL query. 

\hide{
\begin{table}[htbp]
    \centering
    \scalebox{0.85}{
    \begin{tabular}{l|ccc}
    \toprule
    Model   & TR  &  Set-EM\textsubscript{Tran} & Set-EM\textsubscript{All} \\ \midrule
    PGNet & 100.0 & 50.1 & 29.8 \\ 
    PGNet+SE  &  100.0  & \textbf{52.7} & 33.4 \\ 
    PGNet+SE+UTran &  57.3  & 50.4 & \textbf{49.9} \\ \bottomrule
    \end{tabular}
    }
    \caption{Experimental results on \spiderut(\%). ``Set-EM'' is the exact set match accuracy with values. Metric with subscript ``Tran'' shows the results only on translatable data, ``All'' is the metric on the all data, TR is the translate rate.}
    \label{table:spider_exp}
\end{table}
}

\hide{
Currently, there is no model that reports the performance on exact set match with value selection on the Spider leaderborad\footnote{\url{https://yale-lily.github.io/spider}}. Therefore, we compare different settings of our model and give the results in Table~\ref{table:spider_exp}. 
PGNet is the basic pointer-generator network without considering the schema encoding (SE) and untranslatable modular (UTran). PGNet+SE+UTran is the full model we use in our system. 
}

\hide{
with value selection on the translatable examples, illustrated by \emph{Tran} metric. 
The full model with the untranslatable module can achieve 49.9\% accuracy on \spiderut, considering both translatable and untranslatable examples, illustrated by \emph{All} metric. }

%% file: sections/related.tex
\section{Related Work}
\label{sec:related-work}

\paragraph{Natural Language Interfaces to Databases. }
NLIDBs has been studied extensively in the past decades. Thanks to the availability of large-scale datasets~\cite{DBLP:journals/corr/abs-1709-00103,DBLP:conf/acl/RadevKZZFRS18,DBLP:conf/emnlp/YuZYYWLMLYRZR18}, data-driven approaches have dominated the field, in which deep learning based models achieve the best performance in both strongly~\cite{DBLP:journals/corr/abs-1902-01069,DBLP:conf/emnlp/YuYYZWLR19,DBLP:conf/acl/GuoZGXLLZ19} and weakly~\cite{DBLP:conf/acl/LiangBLFL17,DBLP:journals/corr/abs-1909-04849} supervised settings. However, most of existing text-to-SQL datasets include only questions that can be translated into a valid SQL query.
Spider~\cite{DBLP:conf/acl/RadevKZZFRS18} specifically controlled question clarify during data collection to exclude poorly phrased and ambiguous questions. WikiSQL~\cite{DBLP:journals/corr/abs-1709-00103} was constructed on top of manually written synchronous grammars, and the mapping between its questions and SQL queries can be effectively resolved via lexical matching in vector space~\cite{DBLP:journals/corr/abs-1902-01069}.
CoSQL~\cite{DBLP:journals/corr/abs-1909-05378} is by far the only existing corpus to our knowledge which entables data-driven modeling and evaluation of untranslatable question detection. Yet the dataset is of context-dependent nature and contains untranslatable questions of limited variety. We fill in this gap by proposing~\Ours to cover a diverse set of untranslatable user input in text-to-SQL.

\paragraph{Noisy User Input in Semantic Parsing.}
Despite being absent from most large-scale text-to-SQL benchmarks, noisy user input has been frequently encountered and battled with by the semantic parsing community. Underspecification~\cite{archangeli1988aspects} and vagueness~\cite{varzi2001vagueness} have solid linguistic theory foundation. Lexicon-based semantic parsers~\cite{DBLP:conf/uai/ZettlemoyerC05,DBLP:conf/amia/RobertsP17} may reject the input if the lexicon match is unsuccessful. Other approaches for handling untranslatable user input include inference and generating defaults~\cite{DBLP:conf/iui/SetlurTD19}, paraphrasing~\cite{DBLP:journals/tacl/ArthurNSTN15,Arthur2016SemanticPO}, verification~\cite{DBLP:journals/tacl/ArthurNSTN15} and confidence estimation~\cite{DBLP:conf/acl/QuirkLD18}.
We adopt a data-augmentation and discriminative learning based approach, which has demonstrated superior performance in related domains~\cite{DBLP:conf/acl/RajpurkarJL18}

\hide{
\begin{enumerate}
\item Text-to-SQL datasets and methods
\begin{enumerate}
\item WikiSQL, Spider 
\item 
\end{enumerate}
\item Semantic parsing systems
\begin{enumerate}
\item TransX, EUSP
\end{enumerate}
\end{enumerate}
}

%% file: sections/conclusion.tex
\section{Conclusion and Future Work}
\label{sec:conclusion}
We present \Ours, a robust modular cross-domain text-to-SQL system, consisting of semantic parser, untranslatable question detector, human-in-the-loop question corrector, and natural language response generator. 
\Ours has the potential to scale up to hundreds of different domains. It is the first cross-domain text-to-SQL system designed towards industrial applications with rich features, and bridges the demand of sophisticated database analysis and people without any SQL background knowledge. 

The current \Ours system is still a prototype, with very limited user interactions and functions. We will continue to add more features to \Ours, such as voice input, spelling checking, and visualizing the output when appropriate to inspect the translation process. We also plan to improve the performance of core models in \Ours, such as semantic parsing (text-to-SQL), response generation (table-to-text) and context-aware user interaction (text-to-text). A comprehensive evaluation will also be conducted among the users of our system. 

\section*{Acknowledgement}
We thank Dragomir Radev and Tao Yu for valuable research discussions; Melvin Gruesbeck for providing consultancy and frontend design mockups for the demo; Andre Esteva, Yingbo Zhou and Nitish Keskar for advice on model serving; Srinath Reddy Meadusani and Lavanya Karanam for their support on setting up the server infra.

%% file: sections/appendix.tex
\onecolumn
\section{Appendices}
\label{sec:appendices}

\subsection{Construct Untranslatable Questions}
\label{sec:app-ambiguous}

Table~\ref{tab:question-categorization} shows a summary of different types of untranslatable questions based on analysis of CoSQL~\cite{DBLP:journals/corr/abs-1909-05378} and Multi-WOZ~\cite{DBLP:conf/emnlp/BudzianowskiWTC18}.

Table~\ref{tab:transform_example} shows examples of applying question-side and schema-side transformations to convert a translatable question from existing text-to-SQL datasets to an untranslatable question.

\begin{table*}[htpb]
\centering
\includegraphics[width=\textwidth]{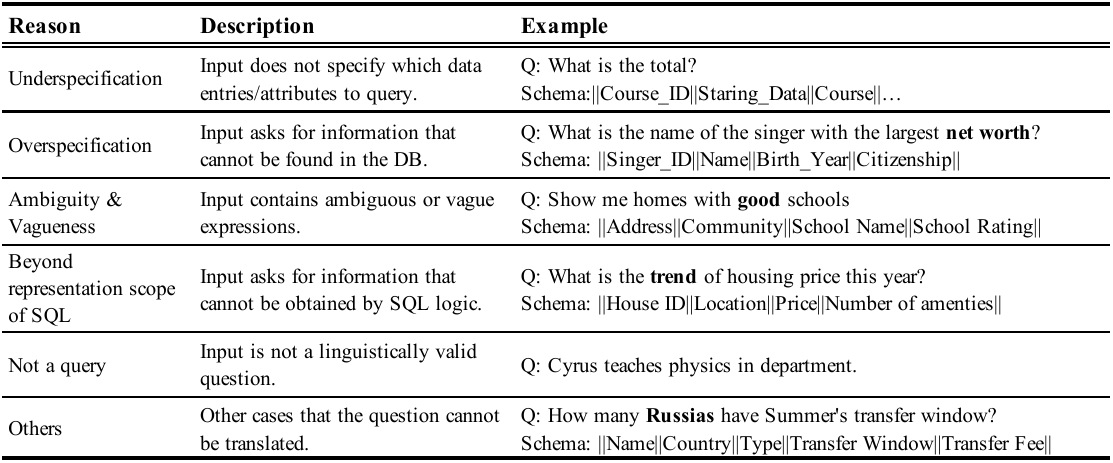}
\caption{Types of untranslatable questions in text-to-SQL identified from manual analysis of CoSQL~\cite{DBLP:journals/corr/abs-1909-05378} and Multi-WOZ~\cite{DBLP:conf/emnlp/BudzianowskiWTC18}. A question span that is problematic for the translation is highlighted when applicable.\label{tab:question-categorization}
}
\end{table*}

\begin{table*}[htpb]
\centering
    \includegraphics[width=0.99 \textwidth]{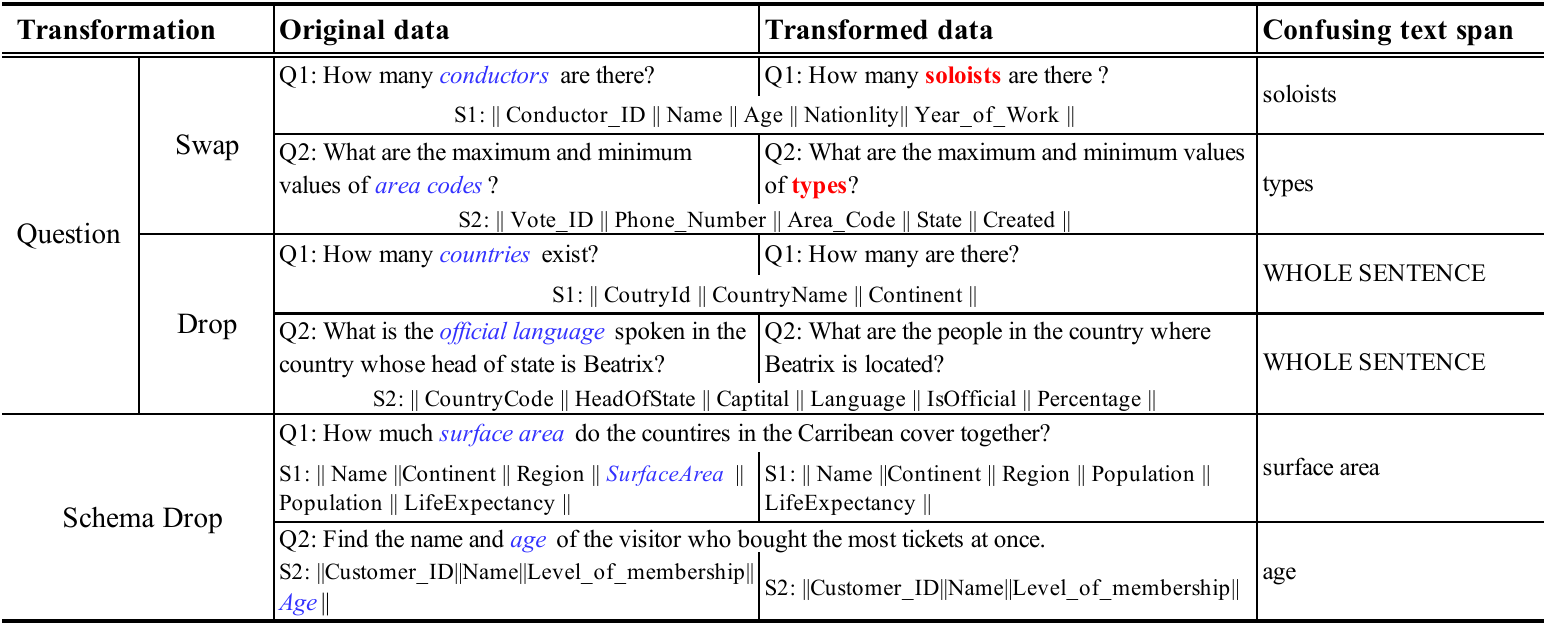}
    \caption{Examples of question-side and schema-side transformations for generating training data for untranslatable question detection. Let $Q$ denote the question and $S$ denote the schema. For each transformation, we provide two examples, i.e., (Q1, S1) and (Q2, S2). The italic and bold fonts highlight phrases before and after transformations. }
    \label{tab:transform_example}
\end{table*}

%% file: acl2020.bbl
\begin{thebibliography}{47}
\expandafter\ifx\csname natexlab\endcsname\relax\def\natexlab#1{#1}\fi

\bibitem[{Archangeli(1988)}]{archangeli1988aspects}
Diana Archangeli. 1988.
\newblock Aspects of underspecification theory.
\newblock \emph{Phonology}, 5(2):183--207.

\bibitem[{Arthur et~al.(2016)Arthur, Neubig, Sakti, and
  Nakamura}]{Arthur2016SemanticPO}
Philip Arthur, Graham Neubig, Sakriani Sakti, and Satoshi Nakamura. 2016.
\newblock Semantic parsing of ambiguous input using multi synchronous grammars.

\bibitem[{Arthur et~al.(2015)Arthur, Neubig, Sakti, Toda, and
  Nakamura}]{DBLP:journals/tacl/ArthurNSTN15}
Philip Arthur, Graham Neubig, Sakriani Sakti, Tomoki Toda, and Satoshi
  Nakamura. 2015.
\newblock \href
  {https://tacl2013.cs.columbia.edu/ojs/index.php/tacl/article/view/654}
  {Semantic parsing of ambiguous input through paraphrasing and verification}.
\newblock \emph{{TACL}}, 3:571--584.

\bibitem[{Bertomeu et~al.(2006)Bertomeu, Uszkoreit, Frank, Krieger, and
  J{\"o}rg}]{bertomeu-etal-2006-contextual}
N{\'u}ria Bertomeu, Hans Uszkoreit, Anette Frank, Hans-Ulrich Krieger, and
  Brigitte J{\"o}rg. 2006.
\newblock \href {https://www.aclweb.org/anthology/W06-3001} {Contextual
  phenomena and thematic relations in database {QA} dialogues: results from a
  wizard-of-{O}z experiment}.
\newblock In \emph{Proceedings of the Interactive Question Answering Workshop
  at {HLT}-{NAACL} 2006}, pages 1--8, New York, NY, USA. Association for
  Computational Linguistics.

\bibitem[{Bogin et~al.(2019{\natexlab{a}})Bogin, Berant, and
  Gardner}]{DBLP:conf/acl/BoginBG19}
Ben Bogin, Jonathan Berant, and Matt Gardner. 2019{\natexlab{a}}.
\newblock Representing schema structure with graph neural networks for
  text-to-sql parsing.
\newblock In \emph{Proceedings of the 57th Conference of the Association for
  Computational Linguistics, {ACL} 2019, Florence, Italy, July 28- August 2,
  2019, Vol. 1}, pages 4560--4565.

\bibitem[{Bogin et~al.(2019{\natexlab{b}})Bogin, Gardner, and
  Berant}]{DBLP:journals/corr/abs-1908-11214}
Ben Bogin, Matt Gardner, and Jonathan Berant. 2019{\natexlab{b}}.
\newblock \href {http://arxiv.org/abs/1908.11214} {Global reasoning over
  database structures for text-to-sql parsing}.
\newblock \emph{CoRR}, abs/1908.11214.

\bibitem[{Budzianowski et~al.(2018)Budzianowski, Wen, Tseng, Casanueva, Ultes,
  Ramadan, and Gasic}]{DBLP:conf/emnlp/BudzianowskiWTC18}
Pawel Budzianowski, Tsung{-}Hsien Wen, Bo{-}Hsiang Tseng, I{\~{n}}igo
  Casanueva, Stefan Ultes, Osman Ramadan, and Milica Gasic. 2018.
\newblock Multiwoz - {A} large-scale multi-domain wizard-of-oz dataset for
  task-oriented dialogue modelling.
\newblock In \emph{Proceedings of the 2018 Conference on Empirical Methods in
  Natural Language Processing, Brussels, Belgium, October 31 - November 4,
  2018}, pages 5016--5026.

\bibitem[{Choi et~al.(2020)Choi, Shin, Kim, and
  Shin}]{DBLP:journals/corr/abs-2004-03125}
DongHyun Choi, Myeong~Cheol Shin, EungGyun Kim, and Dong~Ryeol Shin. 2020.
\newblock {RYANSQL:} recursively applying sketch-based slot fillings for
  complex text-to-sql in cross-domain databases.
\newblock \emph{CoRR}, abs/2004.03125.

\bibitem[{Dahl et~al.(1994)Dahl, Bates, Brown, Fisher, Hunicke{-}Smith,
  Pallett, Pao, Rudnicky, and Shriberg}]{DBLP:conf/naacl/DahlBBFHPPRS94}
Deborah~A. Dahl, Madeleine Bates, Michael Brown, William~M. Fisher, Kate
  Hunicke{-}Smith, David~S. Pallett, Christine Pao, Alexander~I. Rudnicky, and
  Elizabeth Shriberg. 1994.
\newblock Expanding the scope of the {ATIS} task: The {ATIS-3} corpus.
\newblock In \emph{Human Language Technology, Proceedings of a Workshop held at
  Plainsboro, New Jerey, USA, March 8-11, 1994}.

\bibitem[{Devlin et~al.(2019)Devlin, Chang, Lee, and
  Toutanova}]{DBLP:conf/naacl/DevlinCLT19}
Jacob Devlin, Ming{-}Wei Chang, Kenton Lee, and Kristina Toutanova. 2019.
\newblock {BERT:} pre-training of deep bidirectional transformers for language
  understanding.
\newblock In \emph{Proceedings of the 2019 Conference of the North American
  Chapter of the Association for Computational Linguistics: Human Language
  Technologies, {NAACL-HLT} 2019, Minneapolis, MN, USA, June 2-7, 2019, Volume
  1}, pages 4171--4186.

\bibitem[{Dong and Lapata(2016)}]{DBLP:conf/acl/DongL16}
Li~Dong and Mirella Lapata. 2016.
\newblock \href {https://www.aclweb.org/anthology/P16-1004/} {Language to
  logical form with neural attention}.
\newblock In \emph{Proceedings of the 54th Annual Meeting of the Association
  for Computational Linguistics, {ACL} 2016, August 7-12, 2016, Berlin,
  Germany, Volume 1: Long Papers}. The Association for Computer Linguistics.

\bibitem[{Dong and Lapata(2018)}]{DBLP:conf/acl/LapataD18}
Li~Dong and Mirella Lapata. 2018.
\newblock Coarse-to-fine decoding for neural semantic parsing.
\newblock In \emph{Proceedings of the 56th Annual Meeting of the Association
  for Computational Linguistics, {ACL} 2018, Melbourne, Australia, July 15-20,
  2018, Volume 1: Long Papers}, pages 731--742.

\bibitem[{Dong et~al.(2018)Dong, Quirk, and Lapata}]{DBLP:conf/acl/QuirkLD18}
Li~Dong, Chris Quirk, and Mirella Lapata. 2018.
\newblock \href {https://doi.org/10.18653/v1/P18-1069} {Confidence modeling for
  neural semantic parsing}.
\newblock In  \cite{DBLP:conf/acl/2018-1}, pages 743--753.

\bibitem[{Finegan{-}Dollak et~al.(2018)Finegan{-}Dollak, Kummerfeld, Zhang,
  Ramanathan, Sadasivam, Zhang, and Radev}]{DBLP:conf/acl/RadevKZZFRS18}
Catherine Finegan{-}Dollak, Jonathan~K. Kummerfeld, Li~Zhang, Karthik
  Ramanathan, Sesh Sadasivam, Rui Zhang, and Dragomir~R. Radev. 2018.
\newblock \href {https://doi.org/10.18653/v1/P18-1033} {Improving text-to-sql
  evaluation methodology}.
\newblock In  \cite{DBLP:conf/acl/2018-1}, pages 351--360.

\bibitem[{Guo et~al.(2019)Guo, Zhan, Gao, Xiao, Lou, Liu, and
  Zhang}]{DBLP:conf/acl/GuoZGXLLZ19}
Jiaqi Guo, Zecheng Zhan, Yan Gao, Yan Xiao, Jian{-}Guang Lou, Ting Liu, and
  Dongmei Zhang. 2019.
\newblock Towards complex text-to-sql in cross-domain database with
  intermediate representation.
\newblock In \emph{Proceedings of the 57th Conference of the Association for
  Computational Linguistics, {ACL} 2019, Florence, Italy, July 28- August 2,
  2019, Volume 1: Long Papers}, pages 4524--4535.

\bibitem[{Gurevych and Miyao(2018)}]{DBLP:conf/acl/2018-1}
Iryna Gurevych and Yusuke Miyao, editors. 2018.
\newblock \href {https://www.aclweb.org/anthology/volumes/P18-1/}
  {\emph{Proceedings of the 56th Annual Meeting of the Association for
  Computational Linguistics, {ACL} 2018, Melbourne, Australia, July 15-20,
  2018, Volume 1: Long Papers}}. Association for Computational Linguistics.

\bibitem[{Hemphill et~al.(1990)Hemphill, Godfrey, and
  Doddington}]{DBLP:conf/naacl/HemphillGD90}
Charles~T. Hemphill, John~J. Godfrey, and George~R. Doddington. 1990.
\newblock The {ATIS} spoken language systems pilot corpus.
\newblock In \emph{Speech and Natural Language: Proceedings of a Workshop Held
  at Hidden Valley, Pennsylvania, USA, June 24-27, 1990}.

\bibitem[{Hwang et~al.(2019)Hwang, Yim, Park, and
  Seo}]{DBLP:journals/corr/abs-1902-01069}
Wonseok Hwang, Jinyeung Yim, Seunghyun Park, and Minjoon Seo. 2019.
\newblock \href {http://arxiv.org/abs/1902.01069} {A comprehensive exploration
  on wikisql with table-aware word contextualization}.
\newblock \emph{CoRR}, abs/1902.01069.

\bibitem[{Kelkar et~al.(2020)Kelkar, Relan, Bhardwaj, Vaichal, and
  Relan}]{kelkar2020bertrand}
Amol Kelkar, Rohan Relan, Vaishali Bhardwaj, Saurabh Vaichal, and Peter Relan.
  2020.
\newblock Bertrand-dr: Improving text-to-sql using a discriminative re-ranker.
\newblock \emph{arXiv preprint arXiv:2002.00557}.

\bibitem[{Kingma and Ba(2015)}]{DBLP:journals/corr/KingmaB14}
Diederik~P. Kingma and Jimmy Ba. 2015.
\newblock \href {http://arxiv.org/abs/1412.6980} {Adam: {A} method for
  stochastic optimization}.
\newblock In \emph{3rd International Conference on Learning Representations,
  {ICLR} 2015, San Diego, CA, USA, May 7-9, 2015, Conference Track
  Proceedings}.

\bibitem[{Korhonen et~al.(2019)Korhonen, Traum, and
  M{\`{a}}rquez}]{DBLP:conf/acl/2019-1}
Anna Korhonen, David~R. Traum, and Llu{\'{\i}}s M{\`{a}}rquez, editors. 2019.
\newblock \href {https://www.aclweb.org/anthology/volumes/P19-1/}
  {\emph{Proceedings of the 57th Conference of the Association for
  Computational Linguistics, {ACL} 2019, Florence, Italy, July 28- August 2,
  2019, Volume 1: Long Papers}}. Association for Computational Linguistics.

\bibitem[{Li and Jagadish(2014)}]{DBLP:conf/sigmod/LiJ14}
Fei Li and Hosagrahar~Visvesvaraya Jagadish. 2014.
\newblock \href {https://doi.org/10.1145/2588555.2594519} {Nalir: an
  interactive natural language interface for querying relational databases}.
\newblock In \emph{International Conference on Management of Data, {SIGMOD}
  2014, Snowbird, UT, USA, June 22-27, 2014}, pages 709--712. {ACM}.

\bibitem[{Liang et~al.(2017)Liang, Berant, Le, Forbus, and
  Lao}]{DBLP:conf/acl/LiangBLFL17}
Chen Liang, Jonathan Berant, Quoc~V. Le, Kenneth~D. Forbus, and Ni~Lao. 2017.
\newblock \href {https://doi.org/10.18653/v1/P17-1003} {Neural symbolic
  machines: Learning semantic parsers on freebase with weak supervision}.
\newblock In \emph{Proceedings of the 55th Annual Meeting of the Association
  for Computational Linguistics, {ACL} 2017, Vancouver, Canada, July 30 -
  August 4, Volume 1: Long Papers}, pages 23--33. Association for Computational
  Linguistics.

\bibitem[{Lin et~al.(2019)Lin, Bogin, Neumann, Berant, and
  Gardner}]{DBLP:journals/corr/abs-1905-13326}
Kevin Lin, Ben Bogin, Mark Neumann, Jonathan Berant, and Matt Gardner. 2019.
\newblock Grammar-based neural text-to-sql generation.
\newblock \emph{CoRR}, abs/1905.13326.

\bibitem[{Liu et~al.(2019)Liu, Luo, Yang, Wu, Chang, and
  Sui}]{DBLP:conf/acl/LiuLYWCS19}
Tianyu Liu, Fuli Luo, Pengcheng Yang, Wei Wu, Baobao Chang, and Zhifang Sui.
  2019.
\newblock \href {https://doi.org/10.18653/v1/p19-1600} {Towards comprehensive
  description generation from factual attribute-value tables}.
\newblock In  \cite{DBLP:conf/acl/2019-1}, pages 5985--5996.

\bibitem[{Min et~al.(2019)Min, Chen, Hajishirzi, and
  Zettlemoyer}]{DBLP:journals/corr/abs-1909-04849}
Sewon Min, Danqi Chen, Hannaneh Hajishirzi, and Luke Zettlemoyer. 2019.
\newblock \href {http://arxiv.org/abs/1909.04849} {A discrete hard {EM}
  approach for weakly supervised question answering}.
\newblock \emph{CoRR}, abs/1909.04849.

\bibitem[{Popescu et~al.(2003)Popescu, Etzioni, and
  Kautz}]{DBLP:conf/iui/PopescuEK03}
Ana{-}Maria Popescu, Oren Etzioni, and Henry~A. Kautz. 2003.
\newblock \href {https://doi.org/10.1145/604045.604070} {Towards a theory of
  natural language interfaces to databases}.
\newblock In \emph{Proceedings of the 8th International Conference on
  Intelligent User Interfaces, {IUI} 2003, Miami, FL, USA, January 12-15,
  2003}, pages 149--157. {ACM}.

\bibitem[{Rajpurkar et~al.(2018)Rajpurkar, Jia, and
  Liang}]{DBLP:conf/acl/RajpurkarJL18}
Pranav Rajpurkar, Robin Jia, and Percy Liang. 2018.
\newblock Know what you don't know: Unanswerable questions for squad.
\newblock In \emph{Proceedings of the 56th Annual Meeting of the Association
  for Computational Linguistics, {ACL} 2018, Melbourne, Australia, July 15-20,
  2018, Volume 2: Short Papers}, pages 784--789.

\bibitem[{Rajpurkar et~al.(2016)Rajpurkar, Zhang, Lopyrev, and
  Liang}]{Rajpurkar2016SQuAD10}
Pranav Rajpurkar, Jian Zhang, Konstantin Lopyrev, and Percy Liang. 2016.
\newblock Squad: 100, 000+ questions for machine comprehension of text.
\newblock \emph{ArXiv}, abs/1606.05250.

\bibitem[{Roberts and Patra(2017)}]{DBLP:conf/amia/RobertsP17}
Kirk Roberts and Braja~Gopal Patra. 2017.
\newblock \href
  {http://knowledge.amia.org/65881-amia-1.3897810/t003-1.3901461/f003-1.3901462/2729012-1.3901611/2731640-1.3901608}
  {A semantic parsing method for mapping clinical questions to logical forms}.
\newblock In \emph{{AMIA} 2017, American Medical Informatics Association Annual
  Symposium, Washington, DC, USA, November 4-8, 2017}. {AMIA}.

\bibitem[{See et~al.(2017{\natexlab{a}})See, Liu, and
  Manning}]{DBLP:conf/acl/SeeLM17}
Abigail See, Peter~J. Liu, and Christopher~D. Manning. 2017{\natexlab{a}}.
\newblock Get to the point: Summarization with pointer-generator networks.
\newblock In \emph{Proceedings of the 55th Annual Meeting of the Association
  for Computational Linguistics, {ACL} 2017, Vancouver, Canada, July 30 -
  August 4, Volume 1: Long Papers}, pages 1073--1083.

\bibitem[{See et~al.(2017{\natexlab{b}})See, Liu, and
  Manning}]{see-etal-2017-get}
Abigail See, Peter~J. Liu, and Christopher~D. Manning. 2017{\natexlab{b}}.
\newblock \href {https://doi.org/10.18653/v1/P17-1099} {Get to the point:
  Summarization with pointer-generator networks}.
\newblock In \emph{Proceedings of the 55th Annual Meeting of the Association
  for Computational Linguistics (Volume 1: Long Papers)}, pages 1073--1083,
  Vancouver, Canada. Association for Computational Linguistics.

\bibitem[{Setlur et~al.(2016)Setlur, Battersby, Tory, Gossweiler, and
  Chang}]{DBLP:conf/uist/SetlurBTGC16}
Vidya Setlur, Sarah~E. Battersby, Melanie Tory, Rich Gossweiler, and Angel~X.
  Chang. 2016.
\newblock \href {https://doi.org/10.1145/2984511.2984588} {Eviza: {A} natural
  language interface for visual analysis}.
\newblock In \emph{Proceedings of the 29th Annual Symposium on User Interface
  Software and Technology, {UIST} 2016, Tokyo, Japan, October 16-19, 2016},
  pages 365--377. {ACM}.

\bibitem[{Setlur et~al.(2019)Setlur, Tory, and
  Djalali}]{DBLP:conf/iui/SetlurTD19}
Vidya Setlur, Melanie Tory, and Alex Djalali. 2019.
\newblock \href {https://doi.org/10.1145/3301275.3302270} {Inferencing
  underspecified natural language utterances in visual analysis}.
\newblock In \emph{Proceedings of the 24th International Conference on
  Intelligent User Interfaces, {IUI} 2019, Marina del Ray, CA, USA, March
  17-20, 2019}, pages 40--51. {ACM}.

\bibitem[{Sutskever et~al.(2014)Sutskever, Vinyals, and
  Le}]{DBLP:conf/nips/SutskeverVL14}
Ilya Sutskever, Oriol Vinyals, and Quoc~V. Le. 2014.
\newblock Sequence to sequence learning with neural networks.
\newblock In \emph{Advances in Neural Information Processing Systems 27: Annual
  Conference on Neural Information Processing Systems 2014, December 8-13 2014,
  Montreal, Quebec, Canada}, pages 3104--3112.

\bibitem[{Varzi(2001)}]{varzi2001vagueness}
Achille~C Varzi. 2001.
\newblock Vagueness, logic, and ontology.

\bibitem[{Wang et~al.(2018)Wang, Huang, Polozov, Brockschmidt, and
  Singh}]{DBLP:journals/corr/abs-1807-03100}
Chenglong Wang, Po{-}Sen Huang, Alex Polozov, Marc Brockschmidt, and Rishabh
  Singh. 2018.
\newblock \href {http://arxiv.org/abs/1807.03100} {Execution-guided neural
  program decoding}.
\newblock \emph{CoRR}, abs/1807.03100.

\bibitem[{Wolf et~al.(2019)Wolf, Debut, Sanh, Chaumond, Delangue, Moi, Cistac,
  Rault, Louf, Funtowicz, and Brew}]{Wolf2019HuggingFacesTS}
Thomas Wolf, Lysandre Debut, Victor Sanh, Julien Chaumond, Clement Delangue,
  Anthony Moi, Pierric Cistac, Tim Rault, R'emi Louf, Morgan Funtowicz, and
  Jamie Brew. 2019.
\newblock Huggingface's transformers: State-of-the-art natural language
  processing.
\newblock \emph{ArXiv}, abs/1910.03771.

\bibitem[{Yao et~al.(2019)Yao, Su, Sun, and Yih}]{DBLP:conf/emnlp/YaoSSY19}
Ziyu Yao, Yu~Su, Huan Sun, and Wen{-}tau Yih. 2019.
\newblock \href {https://doi.org/10.18653/v1/D19-1547} {Model-based interactive
  semantic parsing: {A} unified framework and {A} text-to-sql case study}.
\newblock In \emph{Proceedings of the 2019 Conference on Empirical Methods in
  Natural Language Processing and the 9th International Joint Conference on
  Natural Language Processing, {EMNLP-IJCNLP} 2019, Hong Kong, China, November
  3-7, 2019}, pages 5446--5457. Association for Computational Linguistics.

\bibitem[{Yu et~al.(2019{\natexlab{a}})Yu, Zhang, Er, Li, Xue, Pang, Lin, Tan,
  Shi, Li, Jiang, Yasunaga, Shim, Chen, Fabbri, Li, Chen, Zhang, Dixit, Zhang,
  Xiong, Socher, Lasecki, and Radev}]{DBLP:journals/corr/abs-1909-05378}
Tao Yu, Rui Zhang, Heyang Er, Suyi Li, Eric Xue, Bo~Pang, Xi~Victoria Lin,
  Yi~Chern Tan, Tianze Shi, Zihan Li, Youxuan Jiang, Michihiro Yasunaga,
  Sungrok Shim, Tao Chen, Alexander~Richard Fabbri, Zifan Li, Luyao Chen, Yuwen
  Zhang, Shreya Dixit, Vincent Zhang, Caiming Xiong, Richard Socher, Walter~S.
  Lasecki, and Dragomir~R. Radev. 2019{\natexlab{a}}.
\newblock \href {http://arxiv.org/abs/1909.05378} {Cosql: {A} conversational
  text-to-sql challenge towards cross-domain natural language interfaces to
  databases}.
\newblock \emph{CoRR}, abs/1909.05378.

\bibitem[{Yu et~al.(2018)Yu, Zhang, Yang, Yasunaga, Wang, Li, Ma, Li, Yao,
  Roman, Zhang, and Radev}]{DBLP:conf/emnlp/YuZYYWLMLYRZR18}
Tao Yu, Rui Zhang, Kai Yang, Michihiro Yasunaga, Dongxu Wang, Zifan Li, James
  Ma, Irene Li, Qingning Yao, Shanelle Roman, Zilin Zhang, and Dragomir~R.
  Radev. 2018.
\newblock Spider: {A} large-scale human-labeled dataset for complex and
  cross-domain semantic parsing and text-to-sql task.
\newblock In \emph{Proceedings of the 2018 Conference on Empirical Methods in
  Natural Language Processing, Brussels, Belgium, Oct 31 - Nov 4, 2018}, pages
  3911--3921.

\bibitem[{Yu et~al.(2019{\natexlab{b}})Yu, Zhang, Yasunaga, Tan, Lin, Li, Er,
  Li, Pang, Chen, Ji, Dixit, Proctor, Shim, Kraft, Zhang, Xiong, Socher, and
  Radev}]{DBLP:conf/acl/YuZYTLLELPCJDPS19}
Tao Yu, Rui Zhang, Michihiro Yasunaga, Yi~Chern Tan, Xi~Victoria Lin, Suyi Li,
  Heyang Er, Irene Li, Bo~Pang, Tao Chen, Emily Ji, Shreya Dixit, David
  Proctor, Sungrok Shim, Jonathan Kraft, Vincent Zhang, Caiming Xiong, Richard
  Socher, and Dragomir~R. Radev. 2019{\natexlab{b}}.
\newblock \href {https://www.aclweb.org/anthology/P19-1443/} {Sparc:
  Cross-domain semantic parsing in context}.
\newblock In  \cite{DBLP:conf/acl/2019-1}, pages 4511--4523.

\bibitem[{Zelle and Mooney(1996)}]{DBLP:conf/aaai/ZelleM96}
John~M. Zelle and Raymond~J. Mooney. 1996.
\newblock Learning to parse database queries using inductive logic programming.
\newblock In \emph{Proceedings of the Thirteenth National Conference on
  Artificial Intelligence and Eighth Innovative Applications of Artificial
  Intelligence Conference, {AAAI} 96, {IAAI} 96, Portland, Oregon, USA, August
  4-8, 1996, Volume 2.}, pages 1050--1055.

\bibitem[{Zellers et~al.(2018)Zellers, Bisk, Schwartz, and
  Choi}]{DBLP:conf/emnlp/ZellersBSC18}
Rowan Zellers, Yonatan Bisk, Roy Schwartz, and Yejin Choi. 2018.
\newblock \href {https://www.aclweb.org/anthology/D18-1009/} {{SWAG:} {A}
  large-scale adversarial dataset for grounded commonsense inference}.
\newblock In \emph{Proceedings of the 2018 Conference on Empirical Methods in
  Natural Language Processing, Brussels, Belgium, October 31 - November 4,
  2018}, pages 93--104. Association for Computational Linguistics.

\bibitem[{Zettlemoyer and Collins(2005)}]{DBLP:conf/uai/ZettlemoyerC05}
Luke~S. Zettlemoyer and Michael Collins. 2005.
\newblock \href
  {https://dslpitt.org/uai/displayArticleDetails.jsp?mmnu=1\&smnu=2\&article\_id=1209\&proceeding\_id=21}
  {Learning to map sentences to logical form: Structured classification with
  probabilistic categorial grammars}.
\newblock In \emph{{UAI} '05, Proceedings of the 21st Conference in Uncertainty
  in Artificial Intelligence, Edinburgh, Scotland, July 26-29, 2005}, pages
  658--666. {AUAI} Press.

\bibitem[{Zhang et~al.(2019)Zhang, Yu, Er, Shim, Xue, Lin, Shi, Xiong, Socher,
  and Radev}]{DBLP:conf/emnlp/YuYYZWLR19}
Rui Zhang, Tao Yu, Heyang Er, Sungrok Shim, Eric Xue, Xi~Victoria Lin, Tianze
  Shi, Caiming Xiong, Richard Socher, and Dragomir~R. Radev. 2019.
\newblock \href {http://arxiv.org/abs/1909.00786} {Editing-based {SQL} query
  generation for cross-domain context-dependent questions}.
\newblock \emph{CoRR}, abs/1909.00786.

\bibitem[{Zhong et~al.(2017)Zhong, Xiong, and
  Socher}]{DBLP:journals/corr/abs-1709-00103}
Victor Zhong, Caiming Xiong, and Richard Socher. 2017.
\newblock Seq2sql: Generating structured queries from natural language using
  reinforcement learning.
\newblock \emph{CoRR}, abs/1709.00103.

\end{thebibliography}
